\documentclass{article}

\usepackage[preprint]{neurips_2025}

\usepackage[utf8]{inputenc} 
\usepackage[T1]{fontenc}    
\usepackage[dvipsnames]{xcolor}         
\usepackage[pagebackref=true,breaklinks=true,letterpaper=true,colorlinks,bookmarks=false, citecolor=ForestGreen]{hyperref}
\usepackage{url}            
\usepackage{booktabs}       
\usepackage{amsfonts}       
\usepackage{nicefrac}       
\usepackage{microtype}      

\definecolor{my_c1}{HTML}{e3f0a3}
\definecolor{my_c2}{HTML}{d0f4e7}

\usepackage[utf8]{inputenc}
\usepackage[T1]{fontenc}
\usepackage[american]{babel}
\usepackage{microtype}

\usepackage{amsmath}
\usepackage{amsthm}
\usepackage{amsfonts}
\usepackage{amssymb}
\usepackage{mathtools}
\usepackage{nicefrac}
\usepackage{algorithm}

\usepackage{graphicx}
\usepackage{subcaption}  
\usepackage{wrapfig}
\usepackage{booktabs}
\usepackage{array}
\usepackage{multirow}
\usepackage{adjustbox}
\usepackage{lscape}
\usepackage{caption}

\usepackage{enumitem}
\usepackage{pifont}
\usepackage{bbding}

\newlength\savewidth

\def\eg{\emph{e.g.}} 
\def\ie{\emph{i.e.}}

\title{Sub-MoE: Efficient Mixture-of-Expert LLMs Compression via Subspace Expert Merging}

\author{
$\text{Lujun Li}^{1 \dag}$,
$\text{Qiyuan Zhu}^{1 \dag}$,
$\text{Jiacheng Wang}^{2}$,
$\textbf{Wei Li}^{3}$, 
$\textbf{Hao Gu}^{1}$,
$\textbf{Sirui Han}^{1 *}$,
$\textbf{Yike Guo}^{1 *}$\\
$^1$Hong Kong University of Science and Technology,
$^2$Xi'an Jiaotong University,\\
$^3$University of Birmingham\\
{\tt\small  \{lliee,qzhuat,siruihan,yikeguo\}@ust.hk}
\thanks{*Corresponding authors, $\dag$ equal contribution. }
}
\vspace{-2.5em}
\begin{document}

\maketitle

\begin{abstract}
Mixture of Experts (MoE) LLMs face significant obstacles due to their massive parameter scale, which imposes memory, storage, and deployment challenges. Although recent expert merging methods promise greater efficiency by consolidating multiple experts, they are fundamentally hindered by parameter conflicts arising from expert specialization. In this paper, we present Sub-MoE, a novel MoE compression framework via Subspace Expert Merging. Our key insight is to perform joint Singular Value Decomposition (SVD) on concatenated expert weights, reducing conflicting parameters by extracting shared $U$-matrices while enabling effective merging of the expert-specific $V$ components.  Specifically, Sub-MoE consists of two innovative phases: (1) Adaptive Expert Clustering, which groups functionally coherent experts via K-means clustering based on cosine similarity of expert outputs; and (2) Subspace Expert Merging, which first enforces Experts Union Decomposition to derive the shared $U$-matrix across experts in the same group, then pursues frequency-based merging for individual $V$-matrices, and finalizes expert reconstruction using the merged $V$-matrix. In this way, we align and fuse experts in a shared subspace, and can be extended with intra-expert compression for further inference optimization. Extensive experiments on Mixtral, DeepSeek, and Qwen-1.5|3 MoE LLMs demonstrate that our Sub-MoE significantly outperforms existing expert pruning and merging methods. Notably, our Sub-MoE maintains 96\%|86\% of original performance with 25\%|50\% expert reduction on Mixtral-8×7B in zero-shot benchmarks. Code will be released at https://github.com/lliai/MoERazor.
\end{abstract}
\section{Introduction}

The Mixture of Experts (MoE) architecture has emerged as a pivotal advancement in Large Language Models (LLMs)~\cite{nguyen2024libmoe},  demonstrated by recent models like DeepSeek-R1~\cite{deepseekai2024deepseekv2strongeconomicalefficient} and Qwen3-MoE~\cite{qwen2.5}.   At its core,  MoE consists of expert networks and a gating mechanism that dynamically routes each input to the most relevant experts. MoE sparsely activates only a small subset of experts, significantly reducing computational costs while scaling model size. However,  MoE LLMs also introduce challenges from their large parameter count, including substantial memory/storage requirements and inference latency that complicate deployment on resource-constrained devices~\cite{song2023powerinfer,li2024dis,li2024als}.  Additionally, distributed implementations face communication~\cite{jiang2024lancet} bottlenecks when synchronizing experts across multiple nodes, impacting real-time performance~\cite{shen2022se,dong2024pruner}.

\begin{figure}[t]
\centering
\includegraphics[width=1.0\linewidth]{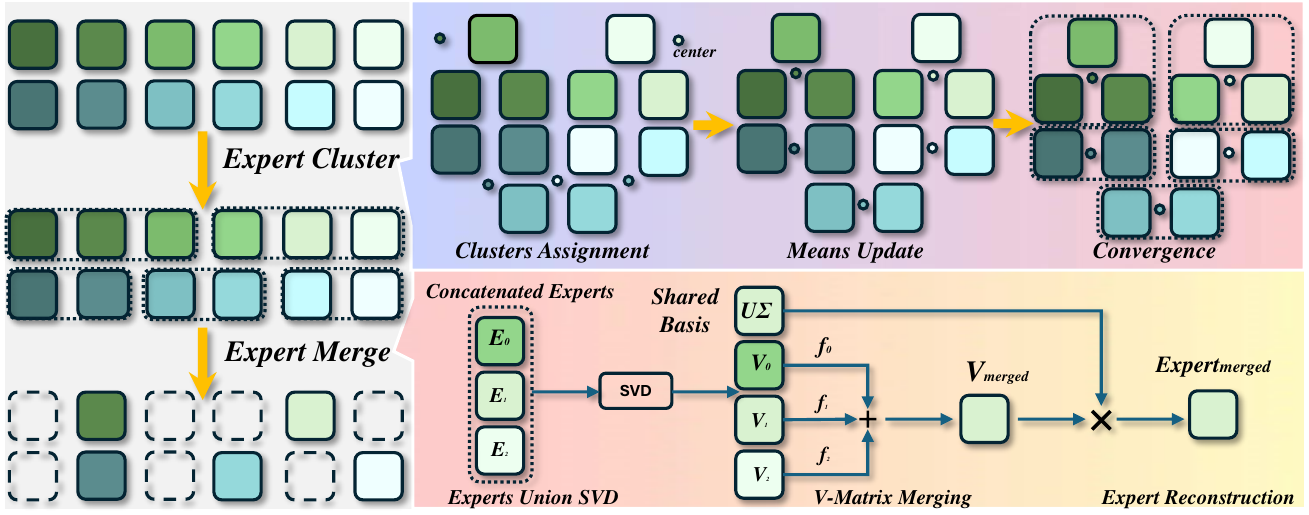}
\caption{Overview of our Sub-MoE framework. The process consists of two main stages: (1) Adaptive Expert Clustering, which groups similar experts via  K-means clustering with steps: Clusters Assignment, Means Update, and Convergence; and (2) Subspace Expert Merging, which aligns and combines experts via Experts Union SVD, $V$-Matrix Merging. and Expert Reconstruction.}
\label{fig:framework}
\vspace{-5mm}
\end{figure}

 To overcome these issues, researchers are developing fundamental yet important expert reduction approaches that can be broadly categorized into two primary approaches: expert pruning and expert merging. Expert pruning methods remove underperforming experts through regularization (\eg, SEER-MoE~\cite{muzio2024seer}) or search-based techniques (\eg, NAEE~\cite{lu2024not}, MoE-I$^2$~\cite{yang2024moe}).  While these approaches effectively reduce parameter counts, they fundamentally discard portions of the model's learned knowledge, resulting in performance degradation  that necessitates resource-intensive fine-tuning to recover.  Expert merging techniques (\eg,  MC-SMoE~\cite{li2024mcsmoe}, HC-SMoE~\cite{chen2024retraining}, and EEP~\cite{liu2024efficient}) propose a promising alternative by preserving knowledge through the consolidation of multiple experts.  However, current merging approaches encounter a critical limitation that undermines their effectiveness: the parameter conflict problem.  This fundamental challenge arises from the core design principle of MoE architectures, where routing mechanisms deliberately create specialized experts with divergent parameter spaces by training them on distinct input distributions. The recent study~\cite{gu2025delta} shows that Mixtral-8×7B demonstrates this divergence, showing inter-expert similarities typically ranging between 0.1$\sim$0.3. When conventional merging operations are applied to such dissimilar experts, catastrophic parameter conflicts emerge that compromise the specialized capabilities of the original experts and significantly degrade overall model performance. Existing merging approaches employ simplistic aggregation functions that cannot effectively reconcile these divergent parameter spaces and often require computationally expensive post-merging operations (\eg, $D^2$-MoE~\cite{gu2025delta}), undermining the efficiency gains. This motivates our core research question: 

\textbf{\textit{(RQ) How can we reduce parameter conflicts among diverse experts and enhance the effectiveness of expert merging?}}

\begin{figure}[t]
\centering
\begin{minipage}[t]{0.6\textwidth}
\centering
\includegraphics[width=1.0\linewidth]{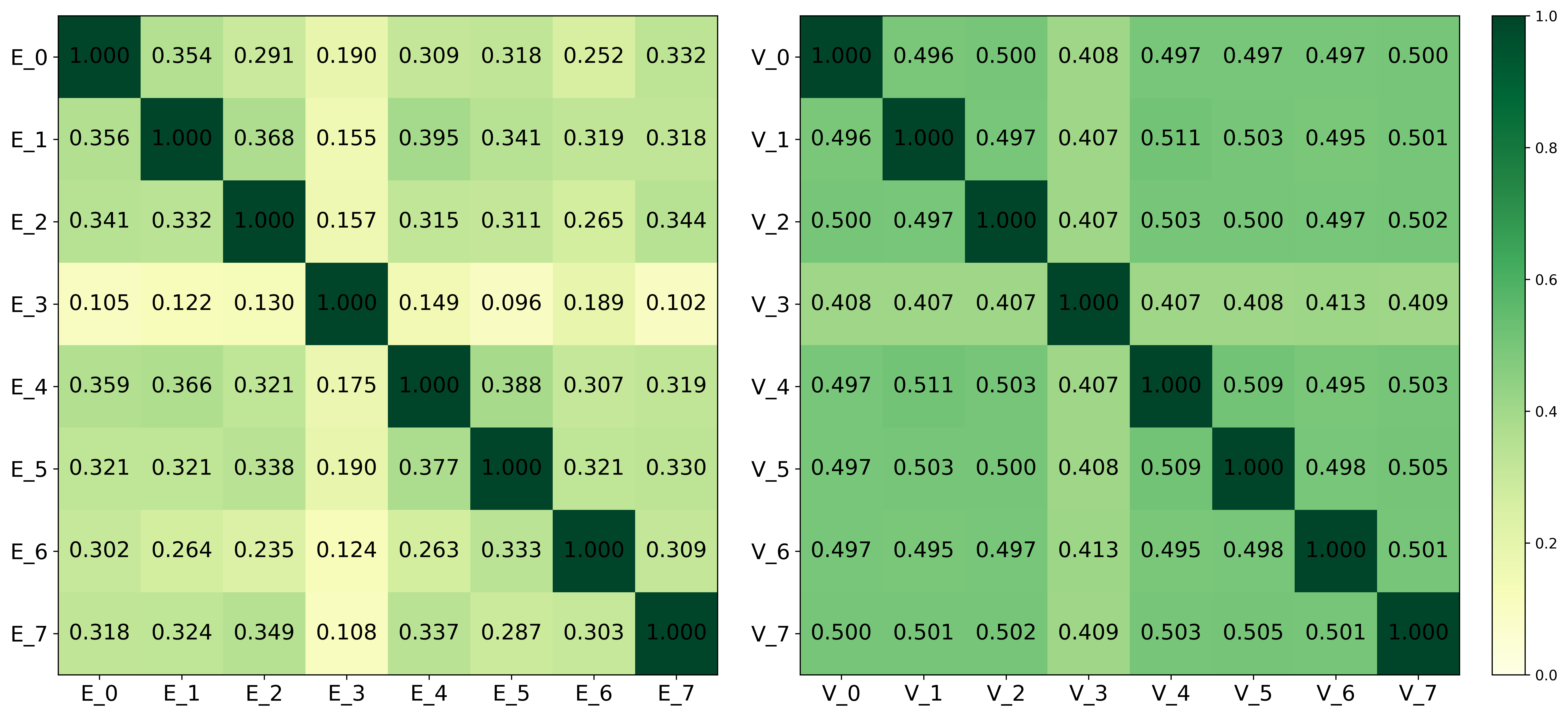}
\caption{Cosine similarity of output of original expert (left) and subspace aligned matrices (right)  on Mixtral-8×7B.}
\label{fig:correlation_heatmaps}
\end{minipage}
\begin{minipage}[t]{0.35\textwidth}
\centering
\includegraphics[width=1.0\linewidth]{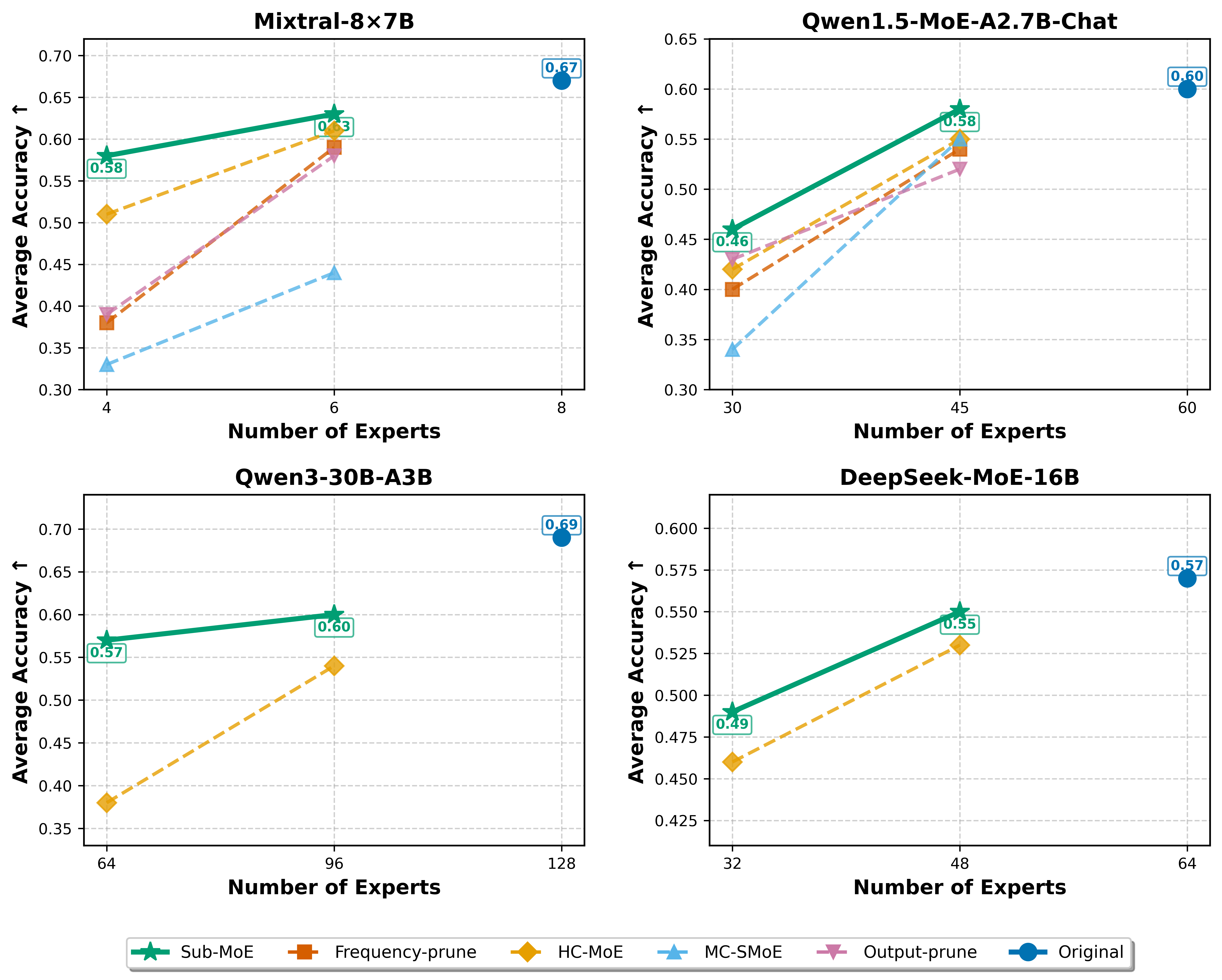}
\caption{Mean accuracy of  MoE compressors on zero-shot tasks.}
\label{fig:moe_expert_comparison}
\end{minipage}
\vspace{-5mm}
\end{figure}

To answer the question, we present Sub-MoE, a novel expert merging framework rooted in subspace-based decomposition and alignment. Our approach leverages Singular Value Decomposition (SVD) to transform the concatenated weight matrices of multiple experts into a shared low-dimensional subspace, represented by a common orthogonal basis $U$, singular values $\Sigma$, and individual projections $V^T$. By performing the merging operation solely on the $V^T$ component—while preserving alignment to the shared $U$—we exploit intrinsic correlations among experts, thereby minimizing conflicting parameters and retaining specialized knowledge. As illustrated in Figure \ref{fig:framework}, our Sub-MoE framework consists of two synergistic stages: Adaptive Expert Clustering and Subspace Expert Merging. In the first stage, we perform adaptive expert grouping via K-means clustering based on output similarities of experts, ensuring that merging is performed on functionally coherent groups.  In addition, we jointly cluster multi-layer experts under a target overall compression ratio and adaptively determine the layer-wise grouping numbers. 
In the second stage,  we concatenate expert weights from the same group and enforce co-decomposition to obtain the shared $U$-matrix across experts and expert-specific $V^T$-matrixs. Figure~\ref{fig:correlation_heatmaps} demonstrates that this subspace-sharing process can align the output of the various experts. For the remaining unmerged components, we further introduce the frequency-based merging strategy that weights expert contributions according to their activation patterns. This approach gives greater influence to frequently activated experts while still preserving capabilities from experts. Finally, the merged weight matrix is reconstructed as $U\Sigma[V_{\text{merged}}]^T$~\footnote{The singular values $\Sigma$ is multiplied in the shared $U$ matrix during the joint SVD process in implementation.}. 
Additionally, we extend Sub-MoE to Sub-MoE† with MoE-specific activation-aware truncated SVD for intra-expert compression for greater parameter efficiency. We incorporate input activation statistics by weighting expert parameters with the whitening matrix of hidden activations, further stabilizing performance at high compression levels.


We conduct comprehensive experiments on  Mixtral 8x7B~\citep{jiang2024mixtral}, Qwen3-235B-A22B~\cite{qwen2.5} Qwen1.5-MoE-A2.7B~\citep{qwen_moe} and DeepSeekMoE-16B-Base~\cite{dai2024deepseekmoe}. As shown in Figure~\ref{fig:moe_expert_comparison}, our proposed Sub-MoE method consistently outperforms existing expert reduction techniques. With Mixtral-8×7B, Sub-MoE maintains 94\% and 87\% of accuracy using only 75\% or 50\% of experts, surpassing HC-SMoE\cite{chen2024retraining} by 13.7\%. For Qwen3 MoE, Sub-MoE maintains 83\% of accuracy with half the experts, while HC-SMoE drops to 55\%. Similarly, on DeepSeek-MoE-16B, our method preserves 86\% of performance with half the experts, outperforming HC-SMoE by 6.5\%. These results affirm the effectiveness and generalizability of our approach across diverse MoE architectures and downstream tasks, positioning Sub-MoE as a principled and scalable solution to expert merging for next-generation MoE LLMs.

\section{Related Work}
\textbf{MoE Compression.} To improve the efficiency of MoE LLMs, researchers have developed numerous system-level optimizations (\eg, expert parallel~\cite{cai2024shortcut} and offloading~\cite{xue2024moe}) and model-level techniques (quantization~\cite{huang2025mixture,dong2024stbllm} and compression~\cite{sarkar2024revisiting}). Among them, expert reduction methods primarily focus on removing redundant experts to achieve optimal efficiency-performance tradeoffs. For optimization-based expert pruning,  SEER-MoE~\cite{muzio2024seer} removes non-professional experts through regularization-based fine-tuning. In search-based expert pruning approaches, NAEE~\cite{lu2024not} trims unimportant experts by minimizing pruning error, while MoE-$I^{2}$~\cite{yang2024moe} employs genetic search strategies. \textbf{In sharp contrast to these pruning approaches, our Sub-MoE explores the merging paradigm that requires neither searching nor fine-tuning.} Other kind methods use weight or hybrid compression for MoE. MoE-Pruner~\cite{xie2024moe} prunes weights based on activations and router logits, STUN~\cite{lee2024stun} combines structured and unstructured pruning and $D^2$-MoE~\cite{gu2025delta} introduces delta compensation. MoE-Compression~\cite{he2024demystifying} provides compressors evaluations. \textbf{Different from these, we highlight that Sub-MoE mainly addresses expert merging rather than weight compression.} 

\textbf{Expert Merging} methods~\cite{li2022branch,zhao-etal-2024-hypermoe} fuse multiple experts into a single one through weighted summation or averaging. For instance, MC-SMoE~\cite{li2024mcsmoe} merges experts with similar routing policies, HC-SMoE~\cite{chen2024retraining} utilizes hierarchical clustering to merge experts in a task-agnostic manner, and EEP~\cite{liu2024efficient} optimizes fusion matrices through evolutionary search algorithms. However, these methods typically employ original weight merging techniques~\cite{wortsman2022model}, which achieve success primarily when handling models with high similarity, such as fine-tuned variants of the same base model~\cite{izmailov2018averaging}. When applied to MoE models with low-similarity experts, these methods generally fail due to significant parameter conflicts during the merging process. \textbf{Our Sub-MoE distinctly differs from previous expert mergers} by addressing this low-similarity issue through expert decomposition into subspaces and enforcing matrices alignment. Thanks to this subspace alignment approach, we can effectively fuse different MoE LLMs without requiring additional training, searching, or complex weight operations.

\section{Methodology}
\label{sec:methodology}

Our Sub-MoE framework consists of two synergistic stages: (1) \textbf{Adaptive Expert Clustering} stage that clusters similar experts, (2) \textbf{Subspace Expert Merging} stage that includes union decomposition, frequency-based $V$-Matrix fusion and reconstruction. The overall process is illustrated in Figure \ref{fig:framework}.

\subsection{Recap of MoE Architecture}
The fundamental principle of MoE models is to dynamically route input data to specialized expert networks. Consider an input token $x \in \mathbb{R}^d$, a set of expert modules $\{E_1, E_2, ..., E_n\}$, and a router network $R$. The output $y$ of an MoE layer is computed as:
\begin{equation}
y = \sum_{i=1}^n G_i(x) \cdot E_i(x), \quad E(x) = (\sigma(x\cdot W_{gate}) \odot (x\cdot W_{up}))\cdot W_{down}
\label{eq:moe_output}
\end{equation}
where $G_i(x)$ represents the routing score for expert $i$, and $E_i(x)$ denotes its output.  Each expert typically implements a feed-forward layer with weight matrices $\{W_{\text{up}}, W_{\text{gate}}  W_{\text{down}}\}$, and  $\sigma$ activation  (\eg,  SiLU function). The router  $R$ employs a top-$k$ strategy with softmax normalization, activating only the most relevant experts for each input token and thereby enhancing computational efficiency.

\subsection{Adaptive Expert Clustering}
A critical challenge in compressing MoE models is identifying which experts can be effectively merged with minimal information loss. Rather than relying on architectural heuristics or arbitrary grouping strategies, we propose a data-driven approach that captures the functional similarity between experts. Our key insight is that experts processing similar input patterns in comparable ways are more amenable to merging. To implement this intuition, we first collect a representative set of input tokens $\mathcal{X} = \{x_1, x_2, ..., x_m\}$ from the target domain. For each expert $E_i$, we compute output vectors across this input set, yielding output collections $\mathcal{Y}_i = \{E_i(x_1), E_i(x_2), ..., E_i(x_m)\}$ that characterize the expert's functional behavior. We then quantify the functional similarity between experts using the average cosine similarity of their outputs:
\begin{equation}
\text{Sim}(E_i, E_j) = \frac{1}{m}\sum_{l=1}^{m}\frac{E_i(x_l) \cdot E_j(x_l)}{||E_i(x_l)|| \cdot ||E_j(x_l)||},
\label{eq:expert_sim}
\end{equation} This similarity metric captures how consistently two experts respond to the same inputs, regardless of their internal parameter representations. Experts with high similarity scores are likely to serve overlapping functions within the model and thus become strong candidates for merging.

Based on this similarity measure, we employ K-means clustering to organize the experts into $k$ coherent groups. This process consists of four key steps:

\textbf{1. Means Initialization:} Initial cluster centroids $C = \{C_1, C_2, ..., C_k\}$ are established through advanced seeding method (\ie, k-means++~\cite{kmeans}) to ensure diverse starting points across the expert functional space.

\textbf{2. Clusters Assignment:} Each expert $E_j$ is assigned to the nearest cluster centroid based on the similarity metric in Equation \ref{eq:expert_sim}, forming expert groups $Q_i$ that share functional characteristics.

\textbf{3. Means Update:} Cluster centroids are recalculated as the mean of all experts assigned to that cluster: $C_i = \frac{1}{|Q_i|}\sum_{E_j \in Q_i}\mathcal{Y}_j$.

\textbf{4. Convergence:} Steps 2 and 3 are repeated until cluster assignments stabilize or maximum iterations are reached, minimizing the objective function:
\begin{equation}
J = \sum_{i=1}^{k}\sum_{E_j \in Q_i}||\mathcal{Y}_j - C_i||^2,
\label{eq:kmeans}
\end{equation}
where $Q_i$ represents the set of experts assigned to cluster $i$. This data-driven approach discovers inherent functional relationships between experts that might not be apparent from architecture alone.

\textbf{Multi-layer Adaptive Allocation:}
Unlike traditional manners that impose uniform reduction across all MoE layers, we introduce a multi-layer adaptive allocation that optimizes the numbers of groups on a per-layer basis. We recognize that different layers within a model exhibit varying degrees of functional redundancy and specialization. By jointly clustering experts on multiple MoE layers while maintaining a target overall compression ratio, our automated clustering process dynamically adjusts clustering centers and determines the optimal number of clusters for each layer without manual intervention. Layers with higher expert similarity naturally form fewer, more cohesive clusters, while those with more diverse patterns maintain more clusters to preserve their specialized capabilities. 

\subsection{Subspace Expert Merging}

\textbf{Problems in Vanilla Expert Merging.}
The central challenge in merging expert networks lies in their different parametric representations. Even when experts serve similar functions, their internal parameters often operate in distinct representation spaces, making direct merging problematic and leading to performance degradation. Given $n$ expert weight matrices $W^{(1)}, W^{(2)}, ..., W^{(n)} \in \mathbb{R}^{O \times I}$, conventional merging methods apply operations directly:

\begin{equation}
W_{\text{merged}} = \sum_{i=1}^{n} \alpha_i W^{(i)},
\label{eq:direct_merge}
\end{equation}
where $\alpha_i$ are weight coefficients. This approach often leads to parameter conflicts because each $W^{(i)}$ operates in its own representation space.


\textbf{Subspace Alignment via Experts Union Decomposition.} We address this challenge by transforming experts into a common subspace before merging. For each expert group identified in the clustering step, we concatenate their weight matrices vertically and apply SVD:
\begin{equation}
\text{SVD}\left([W^{(1)}; W^{(2)}; \dots; W^{(n)}]\right) = U\Sigma [V^{(1)}; V^{(2)}; \dots; V^{(n)}]^T,
\label{eq:svd_concat}
\end{equation}
where $U \in \mathbb{R}^{O \times r}$ contains left singular vectors, which form an orthonormal basis for the input space, $\Sigma \in \mathbb{R}^{r \times r}$ is a diagonal matrix of singular values, and $V \in \mathbb{R}^{r  \times nI}$ contains right singular vectors, which can be partitioned into n blocks, each corresponding to an expert. 

\textbf{Frequency-based $V$-Matrix Merging.} Our method introduces a simple yet effective merging approach that respects the usage patterns of experts in real-world scenarios. We observe that not all experts contribute equally to model outputs--some experts specialize in handling common patterns while others focus on rare cases. Incorporating this frequency information helps preserve the model's capabilities across diverse inputs. For each expert $i$, we calculate its sampling frequency based on actual router activations:
\begin{equation}
f(V_i) = \frac{\sum_{x \in \mathcal{X}}\mathbb{I}[i \in \text{TopK}(G(x), k)]}{|\mathcal{X}|},
\label{eq:freq}
\end{equation}
where $\mathcal{X}$ represents the set of input tokens, $\mathbb{I}[\cdot]$ is the indicator function that equals 1 when expert $i$ is among the top-$k$ experts selected by the routing mechanism for input $x$, and 0 otherwise. This frequency metric captures how often each expert is activated across a representative dataset. We then compute the merged $V$ matrix in each group as a frequency-weighted average:

\begin{equation}
V_{\text{merged}} = \frac{\sum_{i \in Q} f(V_i) \cdot V_i}{\sum_{i \in Q} f(V_i)}
\label{eq:vmerge}
\end{equation}
This frequency-based merging effectively integrates expert information according to their practical utilization patterns, giving greater weight to frequently activated experts while still preserving capabilities from experts.

\textbf{Expert Reconstruction.}  The final merged expert weights are constructed as:

\begin{equation}
W_{\text{merged}} = U\Sigma[V_{\text{merged}}]^T
\label{eq:final}
\end{equation} By construction, all experts within a cluster are merged into a single set of parameters $W_{\text{merged}}$, which is reconstructed using the shared orthogonal basis $U$, singular values $\Sigma$, and the frequency-weighted merged right singular vectors $V_{\text{merged}}$. This process effectively aligns the original experts to a common subspace and compresses them into one representative expert. Through this three-stage process, Sub-MoE achieves effective expert reduction while maintaining performance by operating in a shared subspace, minimizing parameter conflicts, and preserving the essential characteristics of each expert. 

\textbf{Understanding of Subspace Expert Merging.} The foundation of our subspace merging approach can be understood through the lens of manifold learning~\cite{tenenbaum2000global,holiday2019manifold,liu2024pruning} and representation alignment~\cite{singh2020model,tatro2020optimizing}. Each expert can be viewed as operating on a different manifold in parameter space~\cite{vander2021parameter}, but these manifolds often share underlying structures due to the similar functionality of the experts~\cite{gu2025delta}. By applying SVD to the union of experts, we discover a common coordinate system (defined by the left singular vectors $U$) that captures the shared functional subspace across all experts.

The singular values in $\Sigma$ quantify the importance of each dimension in this common subspace, while the right singular vectors in $V$ represent how each expert maps to these dimensions. By sharing the $U$ matrix across all experts within a cluster, we ensure that they operate in the same subspace, making their outputs more compatible and reducing interference when merging. The frequency-weighted merging of the $V$ matrices further ensures that the common subspace is biased toward preserving the functionality of the most frequently used experts, which typically process the most important patterns in the data. Mathematically, this process can also be understood as finding a low-rank approximation of the original expert collection that minimizes the difference between the outputs of the original experts and their reconstructed versions on input data $x$, subject to the constraint that they share a common input transformation $U$. This optimization can be written as:
\begin{equation}
\min_{U, \Sigma, V}  \left\| \sum_{i \in Q}G_i(x) W^{(i)} x - U \Sigma V^T x \right\|_2
\end{equation}
 This formulation reveals why our approach is more effective than direct averaging: we are projecting each expert onto a common subspace before merging, which preserves their essential functionality on real input data while eliminating conflicting parameter representations.

\subsection{Sub-MoE$\dag$ for Intra-Expert Compression}

Sub-MoE reduces expert counts without changes to intra-expert sizes. To further improve the compression ratio for resource-constrained scenarios, we present extended Sub-MoE$\dag$ to reduce the size of $U$, $V$ by truncating before reconstruction.  Beyond previous dense LLM SVD techniques~\cite{wang2024svd}, our Sub-MoE$\dag$ employs MoE-specific activation-aware truncating SVD.

For expert weight matrix $W_i$, we first obtain the activation weighted matrix $S_i$ by measuring the correlation of input activations $X_i$. $S_i$ effectively preserves salience weights and reduces decomposition errors~\cite{wang2024svd}. 
Then, we re-weight each expert’s weight matrix as $W'_i = W_i S_i$. Next, we concatenate the re-weighted weight matrices from all experts in the same group and apply union decomposition:
\begin{equation}
\text{SVD}\left([W'^{(1)}; W'^{(2)}; \dots; W'^{(n)}]\right) = U' \Sigma' [V'^{(1)}; V'^{(2)}; \dots; V'^{(n)}]^T
\end{equation}
 For experts in cluster $Q$, we compute the frequency-weighted merged vector with de-whitening:
\begin{equation}
V_{\text{merged}} = \frac{\sum_{i \in Q} f(V_i) \cdot V'^{(i)} S_i^{-1}}{\sum_{i \in Q} f(V_i)}
\end{equation}
After truncating the smallest singular values in $\Sigma'$ to control the compression ratio, the final merged expert weight is given by:
\begin{equation}
W^{\text{trunc}}_{\text{merged}} = U' \cdot \text{Trunc.}(\Sigma') \cdot V_{\text{merged}}
\end{equation}

This process enables fine-grained control over compression while minimizing information loss, as the activation weighted matrix $S$ enables a direct mapping between singular values and compression loss~\cite{wang2024svd}. By combining expert clustering, subspace alignment, intra-expert compression, and frequency-aware merging, our framework provides a comprehensive solution for MoE model compression across multiple levels of redundancy.

\section{Experiments}
\label{sec:experiments}

In this section, we present a comprehensive evaluation and ablation studies of our Sub-MoE across multiple tasks. All experiments are conducted on eight NVIDIA H800 GPUs.

\subsection{Experimental Setups}
\label{subsec:setup}

\textbf{Models and Datasets.}  We conduct experiments on 4 MoE LLMs: Mixtral 8x7B~\citep{jiang2024mixtral}, Qwen3-235B-A22B~\cite{qwen2.5} Qwen1.5-MoE-A2.7B~\citep{qwen_moe} and DeepSeekMoE-16B-Base~\cite{dai2024deepseekmoe}.  For Mixtral 8x7B, reducing experts from 8 to 4 decreases the model size from 46.7B to 24.2B parameters and reduces computational requirements from 2989 to 1546 GFLOPs. Similarly, for Qwen1.5-MoE, reducing experts from 60 to 30 results in a 43\% reduction in model size (from 14.3B to 8.1B parameters). To evaluate our method comprehensively, we use two types of metrics: (1) perplexity on standard language modeling benchmarks including WikiText-2, PTB, and C4, and (2) accuracy on eight diverse reasoning and understanding tasks~\citep{eval-harness} like ARC~\citep{clark2018arc}, BoolQ~\citep{clark2019boolq}, HellaSwag~\citep{zellers2019hellaswag}, MMLU~\citep{hendrycks2021mmlu}, OBQA ~\citep{OpenBookQA2018}, RTE~\citep{bentivogli2009rte}, and WinoG.~\citep{sakaguchi2019winogrande}.

\textbf{Implementation Details.} For our method, we use a calibration dataset of 128 samples, each containing 2,048 tokens sampled from WikiText-2, unless otherwise specified. In our subspace alignment process, we apply expert grouping based on functional similarity using the expert output metric and K-means clustering as our default configuration. For expert merging, we employ frequency-based V matrix merging, which weighs by their activation frequency in the calibration data. We provide reproduced results of expert pruning methods, frequency-prune, output-prune based on frequency in MoE-Compression~\cite{he2024demystifying} and expert merge methods (\ie, MC-SMoE~\cite{li2024mcsmoe}, HC-SMoE~\cite{chen2024retraining}).

\begin{table}[t]
  \centering
  \caption{Comparisons of expert prune/merge methods in multiple MoE LLMs. We report perplexity (lower is better↓) on language modeling tasks and accuracy (higher is better↑) on reasoning tasks.}
  \label{tab:mixtral_comparison}
  \resizebox{140mm}{!}{
    \begin{tabular}{llrrr|rrrrrrrrr}
    \toprule
    
    Expert & Method & WikiText-2↓ & PTB↓  & C4↓   & ARC-c & ARC-e & BoolQ & HellaS. & MMLU  & OBQA  & RTE   & WinoG. & Average↑ \\
   \midrule
      \multicolumn{14}{c}{\textbf{Mixtral-8×7B}} \\
      
    Num=8 & Original & 3.98  & 14.79 & \multicolumn{1}{r}{7.33} & 0.56  & 0.84  & 0.85  & 0.65  & 0.67  & 0.35  & 0.71  & 0.76  & 0.67 \\
    \midrule
    \multirow{5}[2]{*}{Num=6} & Frequency-prune  & 6.22  & 18.00 & 9.94  & 0.48  & 0.78  & 0.78  & 0.57  & 0.47  & 0.32  & 0.55  & 0.75  & 0.59 \\
          & Output-prune  & 6.17  & 18.28 & 9.63  & 0.47  & 0.77  & 0.75  & 0.58  & 0.46  & 0.30  & 0.60  & 0.75  & 0.58 \\
          & MC-SMoE & 58.11 & 173.51 & 98.86 & 0.29  & 0.60  & 0.59  & 0.43  & 0.25  & 0.20  & 0.53  & 0.60  & 0.44 \\
          & HC-SMoE & 5.92  & 18.70 & 9.49  & 0.45  & 0.73  & 0.83  & 0.57  & 0.56  & 0.29  & 0.69  & 0.75  & 0.61 \\

          &  \textbf{Sub-MoE (Ours) } & \textbf{5.16} & \textbf{18.58} & \textbf{8.54} & \textbf{0.49} & \textbf{0.80} & \textbf{0.86} & \textbf{0.62} & \textbf{0.59} & \textbf{0.32} & \textbf{0.65} & \textbf{0.75} & \textbf{0.64} \\
   \midrule
    \multirow{5}[2]{*}{Num=4} & Frequency-prune  & 17.45 & 79.43 & 22.40 & 0.22  & 0.39  & 0.60  & 0.36  & 0.24  & 0.14  & 0.53  & 0.53  & 0.38 \\
          & \multicolumn{1}{p{7.4em}}{Output-prune } & 15.40 & 81.96 & 20.08 & 0.21  & 0.39  & 0.63  & 0.38  & 0.24  & 0.16  & 0.54  & 0.56  & 0.39 \\
          & MC-SMoE & 854.05 & 1204.41 & 1408.10 & 0.21  & 0.28  & 0.52  & 0.28  & 0.25  & 0.11  & 0.50  & 0.52  & 0.33 \\
          & HC-SMoE & 9.88  & 34.13 & 16.78 & 0.32  & 0.61  & 0.75  & 0.49  & 0.39  & 0.26  & 0.61  & 0.67  & 0.51 \\

          &   \textbf{Sub-MoE (Ours) }  & \textbf{6.97} & \textbf{26.88} & \textbf{10.64} & \textbf{0.45} & \textbf{0.75} & \textbf{0.84} & \textbf{0.57} & \textbf{0.48} & \textbf{0.29} & \textbf{0.57} & \textbf{0.72} & \textbf{0.58} \\
   \midrule
      \midrule
       \multicolumn{14}{c}{\textbf{Qwen1.5-MoE-A2.7B-Chat}} \\
    Num=60 & Original & 8.12  & 12.97 & 11.62 & 0.40  & 0.71  & 0.81  & 0.59  & 0.60  & 0.31  & 0.74  & 0.66  & 0.60 \\
        \midrule
    \multirow{5}[1]{*}{Num=45} & Frequency-prune  & 11.44 & 15.40 & 14.24 & 0.33  & 0.57  & 0.77  & 0.55  & 0.43  & 0.29  & 0.73  & 0.65  & 0.54 \\
   
          & Output-prune  & 11.09 & 18.00 & 16.65 & 0.34  & 0.59  & 0.71  & 0.52  & 0.48  & 0.27  & 0.66  & 0.59  & 0.52 \\
          & MC-SMoE & 12.76 & 17.45 & 16.39 & 0.37  & 0.65  & 0.76  & 0.53  & 0.38  & 0.25  & 0.78  & 0.67  & 0.55 \\
          & HC-SMoE & 11.62 & 16.39 & 15.40 & 0.34  & 0.66  & 0.75  & 0.53  & 0.50  & 0.28  & 0.70  & 0.61  & 0.55 \\

          &       \textbf{Sub-MoE (Ours) }  & \textbf{9.48} & \textbf{14.84} & \textbf{13.16} & \textbf{0.37} & \textbf{0.69} & \textbf{0.80} & \textbf{0.56} & \textbf{0.53} & \textbf{0.30} & \textbf{0.76} & \textbf{0.66} & \textbf{0.58} \\
    \midrule
    \multirow{5}[2]{*}{Num=30} & Frequency-prune  & 32.09 & 42.52 & 39.94 & 0.26  & 0.41  & 0.62  & 0.39  & 0.25  & 0.20  & 0.55  & 0.57  & 0.40 \\
          & Output-prune  & 38.71 & 36.94 & 42.52 & 0.27  & 0.51  & 0.64  & 0.40  & 0.33  & 0.19  & 0.55  & 0.54  & 0.43 \\
          & MC-SMoE & 586.98 & 1865.43 & 2889.24 & 0.19  & 0.33  & 0.57  & 0.29  & 0.23  & 0.18  & 0.45  & 0.52  & 0.34 \\
          & HC-SMoE & 25.60 & 38.18 & 48.94 & 0.25  & 0.50  & 0.64  & 0.33  & 0.35  & 0.19  & 0.50  & 0.57  & 0.42 \\
          &      \textbf{Sub-MoE (Ours) } & \textbf{17.51} & \textbf{29.00} & \textbf{25.28} & \textbf{0.32} & \textbf{0.58} & \textbf{0.51} & \textbf{0.46} & \textbf{0.38} & \textbf{0.25} & \textbf{0.57} & \textbf{0.58} & \textbf{0.46} \\
    \midrule
       \midrule
       \multicolumn{14}{c}{\textbf{Qwen3-30B-A3B}} \\
    Num=128 & Original & 8.64  & 15.40 & 14.47 & 0.53  & 0.80  & 0.89  & 0.60  & 0.78  & 0.35  & 0.83  & 0.71  & 0.69 \\
    \midrule
    \multirow{2}[2]{*}{Num=96} & HC-SMoE & 18.86 & 31.11 & 29.68 & 0.35  & 0.64  & 0.82  & 0.40  & 0.55  & 0.22  & 0.73  & 0.61  & 0.54 \\
          &   \textbf{Sub-MoE (Ours) }  & 13.59 & 23.48 & 21.38 & 0.44  & 0.70  & 0.86  & 0.47  & 0.65  & 0.25  & 0.76  & 0.66  & 0.60 \\
    \midrule
    \multirow{2}[2]{*}{Num=64} & HC-SMoE & 72.33 & 162.99 & 148.41 & 0.23  & 0.44  & 0.63  & 0.29  & 0.30  & 0.13  & 0.50  & 0.50  & 0.38 \\
          &       \textbf{Sub-MoE (Ours) } & 21.05 & 43.19 & 36.37 & 0.40  & 0.68  & 0.84  & 0.41  & 0.56  & 0.23  & 0.77  & 0.63  & 0.57 \\
    \midrule
       \midrule
       \multicolumn{14}{c}{\textbf{DeepSeek-MoE-16B}} \\
    Num=64 & Original & 6.51  & 9.72  & 10.15 & 0.44  & 0.76  & 0.72  & 0.58  & 0.38  & 0.33  & 0.62  & 0.70  & 0.57 \\
   \midrule
    \multirow{2}[2]{*}{Num=48} & HC-SMoE & 9.13  & 12.19 & 13.45 & 0.39  & 0.71  & 0.72  & 0.52  & 0.30  & 0.30  & 0.64  & 0.70  & 0.53 \\
 
          &       \textbf{Sub-MoE (Ours) }  & \textbf{8.48} & \textbf{11.29} & \textbf{12.60} & \textbf{0.40} & \textbf{0.72} & \textbf{0.73} & \textbf{0.54} & \textbf{0.32} & \textbf{0.27} & \textbf{0.66} & \textbf{0.70} & \textbf{0.55} \\
   \midrule
    \multirow{2}[2]{*}{Num=32} & HC-SMoE & 15.34 & 21.07 & 23.30 & 0.31  & 0.60  & 0.69  & 0.43  & 0.24  & 0.20  & 0.57  & 0.64  & 0.46 \\
    
          &       \textbf{Sub-MoE (Ours) } & \textbf{13.71} & \textbf{18.35} & \textbf{20.70} & \textbf{0.32} & \textbf{0.63} & \textbf{0.68} & \textbf{0.44} & \textbf{0.25} & \textbf{0.22} & \textbf{0.65} & \textbf{0.65} & \textbf{0.49} \\
    \bottomrule
    \end{tabular}%
    }
\end{table}%

\subsection{Performance Comparisons} 

Table \ref{tab:mixtral_comparison} presents comprehensive comparisons of our Sub-MoE method against baseline approaches across four different MoE language models with varying degrees of expert reduction. The results demonstrate the consistent superiority of our proposed method across all evaluated models and compression ratios. For Mixtral-8×7B, when compressing from 8 to 6 experts, Sub-MoE achieves significantly better perplexity scores on WikiText-2 (5.16), PTB (18.58), and C4 (8.54) compared to pruning-based methods and other merging approaches. Notably, when reducing to just 4 experts (50\% compression), our approach maintains impressive performance with an average accuracy of 0.58 across reasoning tasks, substantially outperforming the next best method HC-SMoE (0.51) and far surpassing pruning-based approaches that struggle to exceed 0.39 average accuracy. The performance gap becomes even more pronounced with the Qwen1.5-MoE-A2.7B-Chat model, where compressing from 60 to 45 experts shows our Sub-MoE maintaining near-original performance (0.58 vs. 0.60) while other methods show significant degradation. When examining larger models like Qwen3-30B-A3B with 128 experts, Sub-MoE demonstrates remarkable resilience even at 50\% compression (64 experts), maintaining 0.57 average accuracy while HC-SMoE drops dramatically to 0.38. This pattern repeats with DeepSeek-MoE-16B, where our approach consistently preserves more of the original model's capabilities across both language modeling and reasoning tasks. The substantial performance advantage of Sub-MoE becomes increasingly evident as the compression ratio increases, highlighting the effectiveness of our subspace alignment approach in preserving expert functionality compared to traditional merging or pruning techniques. 
\begin{table}[t]
  \centering
  \caption{Performance of Sub-MoE and MC-SMoE under extra intra-expert compression ratios.  Runtime denotes runtime throughput (Tokens/sec) on 8x H800 GPUs.}
  \label{tab:svd_ratio}
    \resizebox{140mm}{!}{
    \begin{tabular}{llllrrr|rrrrrrrrr}
    \toprule
       
    Model & Ratio& Runtime  & Method & WikiText-2↓ & PTB↓  & \multicolumn{1}{r|}{C4↓} & ARC-c & ARC-e & BoolQ & HellaS. & MMLU  & OBQA  & RTE   & WinoG. & Average↑ \\
    \midrule
    Mixtral 8x7B & 0 & 87.7     & Original & 3.98  & 12.99 & 6.78  & 0.56  & 0.84  & 0.85  & 0.65  & 0.67  & 0.35  & 0.71  & 0.76  & 0.67 \\
    \midrule
    \multirow{6}[2]{*}{Mixtral 6x7B} & \multirow{2}[1]{*}{10\%} & \multirow{2}[1]{*}{93.1 }& MC-SMoE & 9.05  & 65.86 & 25.79 & 0.35  & 0.66  & 0.62  & 0.43  & 0.39  & 0.24  & 0.53  & 0.66  & 0.48 \\
           &  &      & \multicolumn{1}{l}{  \textbf{Sub-MoE$\dag$ (Ours) } } & \textbf{6.50} & \textbf{40.33} & \textbf{13.43} & \textbf{0.44} & \textbf{0.75} & \textbf{0.78} & \textbf{0.52} & \textbf{0.52} & \textbf{0.31} & \textbf{0.62} & \textbf{0.72} & \textbf{0.58} \\
       & \multirow{2}[0]{*}{20\%}   & \multirow{2}[0]{*}{104.7}  & MC-SMoE  & 12.96 & 115.58 & 49.71 & 0.26  & 0.54  & 0.62  & 0.36  & 0.33  & 0.18  & 0.53  & 0.60  & 0.43 \\
          &  &     & \multicolumn{1}{l}{  \textbf{Sub-MoE$\dag$ (Ours) } } & \textbf{7.97} & \textbf{63.37} & \textbf{20.48} & \textbf{0.38} & \textbf{0.70} & \textbf{0.67} & \textbf{0.46} & \textbf{0.43} & \textbf{0.28} & \textbf{0.58} & \textbf{0.68} & \textbf{0.52} \\
          & \multirow{2}[1]{*}{30\%} & \multirow{2}[1]{*}{120.9  }  & MC-SMoE  & 50.49 & 314.19 & 135.13 & 0.20  & 0.42  & 0.39  & 0.34  & 0.26  & 0.16  & 0.53  & 0.54  & 0.35 \\
           &  &      & \multicolumn{1}{l}{  \textbf{Sub-MoE$\dag$ (Ours) } } & \textbf{11.22} & \textbf{106.72} & \textbf{38.16} & \textbf{0.29} & \textbf{0.60} & \textbf{0.63} & \textbf{0.38} & \textbf{0.33} & \textbf{0.22} & \textbf{0.53} & \textbf{0.61} & \textbf{0.45} \\
    \midrule
    \multirow{6}[2]{*}{Mixtral 4x7B} & \multirow{2}[1]{*}{10\%} & \multirow{2}[1]{*}{95.3 } & MC-SMoE & 708.03 & 1595.59 & 1204.41 & 0.21  & 0.26  & 0.40  & 0.27  & 0.27  & 0.11  & 0.50  & 0.49  & 0.31 \\
           &  &       & \multicolumn{1}{l}{  \textbf{Sub-MoE$\dag$ (Ours) } } & \textbf{8.60} & \textbf{54.27} & \textbf{16.15} & \textbf{0.40} & \textbf{0.70} & \textbf{0.72} & \textbf{0.48} & \textbf{0.41} & \textbf{0.28} & \textbf{0.56} & \textbf{0.69} & \textbf{0.53} \\
          & \multirow{2}[0]{*}{20\%}   & \multirow{2}[0]{*}{108.2 } &MC-SMoE  & 730.51 & 1698.49 & 1322.79 & 0.21  & 0.25  & 0.40  & 0.27  & 0.26  & 0.12  & 0.52  & 0.50  & 0.32 \\
           &  &        & \multicolumn{1}{l}{  \textbf{Sub-MoE$\dag$ (Ours) } } & \textbf{10.23} & \textbf{83.93} & \textbf{23.71} & \textbf{0.34} & \textbf{0.65} & \textbf{0.65} & \textbf{0.43} & \textbf{0.37} & \textbf{0.23} & \textbf{0.53} & \textbf{0.65} & \textbf{0.48} \\
          & \multirow{2}[1]{*}{30\%} & \multirow{2}[1]{*}{122.7} & MC-SMoE  & 2113.81 & 2630.68 & 2471.30 & 0.22  & 0.27  & 0.44  & 0.26  & 0.25  & 0.12  & 0.47  & 0.51  & 0.32 \\
           &  &        & \multicolumn{1}{l}{  \textbf{Sub-MoE$\dag$ (Ours) } } & \textbf{14.82} & \textbf{147.42} & \textbf{47.70} & \textbf{0.26} & \textbf{0.55} & \textbf{0.62} & \textbf{0.36} & \textbf{0.29} & \textbf{0.19} & \textbf{0.53} & \textbf{0.61} & \textbf{0.43} \\
    \bottomrule
    \end{tabular}%
    }
\end{table}%

\subsection{Effect of Intra-Expert Compression}
Table~\ref{tab:svd_ratio} compares Sub-MoE$\dag$ against MC-SMoE under various intra-expert compression ratios. Our method consistently outperforms MC-SMoE across all settings, with dramatic differences at higher compression rates. When compressing Mixtral 4x7B with a 10\% ratio, Sub-MoE$\dag$ maintains reasonable perplexity scores (8.60, 54.27, 16.15 on WikiText-2, PTB, and C4), while MC-SMoE suffers catastrophic degradation. This performance gap widens as compression increases, demonstrating Sub-MoE$\dag$'s robustness to parameter reduction.  As shown in  Figure~\ref{fig:exp} (Left), Sub-MoE$\dag$ with fine-tuning (Sub-MoE$\dag$+FT) is able to additionally recover the accuracy significantly compared to more compressors, achieving gains of 4-6\% over the base Sub-MoE across benchmarks. Our method obtains a stabilizing gain across diverse reasoning tasks, outperforming the other competitor ($D^2$-MoE~\cite{gu2025delta}) by 6\% on ARC-e, 5\% on WinoGrande, and 6\% on the challenging ARC-c dataset, demonstrating robust generalization capabilities even after substantial compression.

\subsection{Ablation Study}

\begin{table}[t]
  \centering
  \caption{Ablation on our (A) Expert Clustering and  (B) Subspace Merging for Mixtral 8x7B→6x7B. }
  \label{tab:ablation_components}
    \resizebox{140mm}{!}
    {
    \begin{tabular}{llrrr|rrrrrrrrr}
    \toprule
       
    Settings & Options & \multicolumn{1}{c}{WikiText-2↓} & \multicolumn{1}{c}{PTB↓} & \multicolumn{1}{c}{C4↓} & \multicolumn{1}{c}{ARC-c} & \multicolumn{1}{c}{ARC-e} & \multicolumn{1}{c}{BoolQ} & \multicolumn{1}{c}{HellaS.} & \multicolumn{1}{c}{MMLU} & \multicolumn{1}{c}{OBQA} & \multicolumn{1}{c}{RTE} & \multicolumn{1}{c}{WinoG.} & \multicolumn{1}{c}{Average↑} \\
    \midrule
     \multicolumn{14}{c}{\textbf{(A) Adaptive Expert Clustering Settings}} \\
    \multirow{3}[1]{*}{Clustering Layer} & Sub-MoE (1-Layer) & 5.47  & 23.77 & 9.51  & 0.47  & 0.79  & 0.83  & 0.59  & 0.55  & 0.31  & 0.64  & 0.75  & 0.62 \\
          &  \textbf{Sub-MoE (2-Layer)} & \textbf{5.16} & \textbf{18.58} & \textbf{8.54} & \textbf{0.49} & \textbf{0.80} & \textbf{0.86} & \textbf{0.62} & \textbf{0.59} & \textbf{0.32} & \textbf{0.65} & \textbf{0.75} & \textbf{0.64} \\
          & Sub-MoE (3-Layer) & 7.02  & 49.99 & 7.22  & 0.38  & 0.74  & 0.65  & 0.48  & 0.46  & 0.24  & 0.56  & 0.71  & 0.53 \\
    \midrule
    \multirow{3}[0]{*}{Similarity Metric} & Sub-MoE (Router-logits) & 5.65  & 16.59 & 9.17  & 0.48  & 0.76  & 0.82  & 0.61  & 0.59  & 0.31  & 0.64  & 0.74  & 0.62 \\
          & Sub-MoE (Weight) & 5.41  & 22.11 & 8.97  & 0.49  & 0.78  & 0.86  & 0.63  & 0.60  & 0.33  & 0.69  & 0.72  & 0.63 \\
          &  \textbf{Sub-MoE (Expert output)} & \textbf{5.16} & \textbf{18.58} & \textbf{8.54} & \textbf{0.49} & \textbf{0.80} & \textbf{0.86} & \textbf{0.62} & \textbf{0.59} & \textbf{0.32} & \textbf{0.65} & \textbf{0.75} & \textbf{0.64} \\
    \midrule
    \multirow{3}[0]{*}{Clustering Alg.} & Sub-MoE (Random) & 6.19  & 18.70 & 9.87  & 0.50  & 0.75  & 0.80  & 0.57  & 0.56  & 0.28  & 0.61  & 0.71  & 0.60 \\
          & Sub-MoE (Hierarchical) & 5.46  & 19.30 & 9.01  & 0.50  & 0.73  & 0.82  & 0.61  & 0.62  & 0.33  & 0.69  & 0.71  & 0.63 \\
          & \textbf{Sub-MoE (K-means)} & \textbf{5.16} & \textbf{18.58} & \textbf{8.54} & \textbf{0.49} & \textbf{0.80} & \textbf{0.86} & \textbf{0.62} & \textbf{0.59} & \textbf{0.32} & \textbf{0.65} & \textbf{0.75} & \textbf{0.64} \\
              \midrule
                 \midrule
     \multicolumn{14}{c}{\textbf{(B) Subspace Expert Merging Settings}} \\
    \multirow{2}[1]{*}{ $U$-Sharing} & Sub-MoE (Vanilla SVD)& 7.02  & 23.91 & 10.67 & 0.42  & 0.70  & 0.61  & 0.59  & 0.63  & 0.28  & 0.72  & 0.69  & 0.58 \\ 
          &  \textbf{Sub-MoE (Union SVD)} & \textbf{5.16} & \textbf{18.58} & \textbf{8.54} & \textbf{0.49} & \textbf{0.80} & \textbf{0.86} & \textbf{0.62} & \textbf{0.59} & \textbf{0.32} & \textbf{0.65} & \textbf{0.75} & \textbf{0.64} \\
    \midrule
    \multirow{3}[1]{*}{ $V$-Merging} & Sub-MoE (Drop) & 5.53  & 19.77 & 9.05  & 0.50  & 0.80  & 0.84  & 0.59  & 0.59  & 0.32  & 0.61  & 0.71  & 0.61  \\
   & Sub-MoE (Average) & 5.31  & 18.63 & 8.88  & 0.50  & 0.81  & 0.85  & 0.61  & 0.59  & 0.31  & 0.64  & 0.74  & 0.62\\
        & \textbf{Sub-MoE (Frequency)}  & \textbf{5.16} & \textbf{18.58} & \textbf{8.54} & \textbf{0.49} & \textbf{0.80} & \textbf{0.86} & \textbf{0.62} & \textbf{0.59} & \textbf{0.32} & \textbf{0.65} & \textbf{0.75} & \textbf{0.64} \\

    \bottomrule
    \end{tabular}%
    }
    \vspace{-5mm}
\end{table}%

\begin{figure}[t]
\centering
\includegraphics[width=1.0\linewidth]{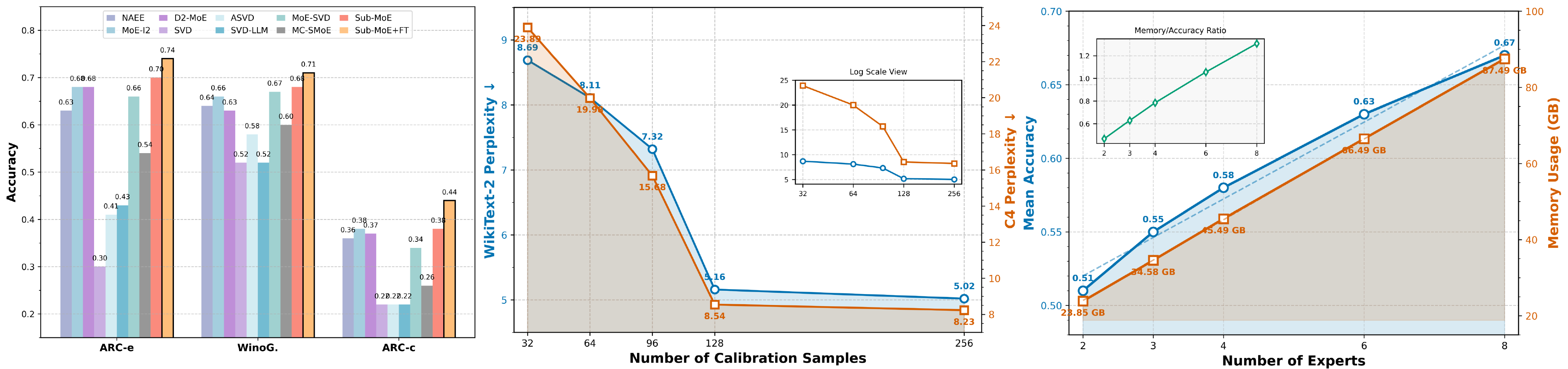}
\vspace{-2mm}
\caption{ \textbf{Left}: Comparison of Sub-MoE$\dag$ against SVD-LLM~\cite{wang2024svd}, NAEE~\cite{lu2024not}, MoE-I$^2$~\cite{yang2024moe}, MoE-SVD~\cite{li2025structured}, $D^2$-MoE~\cite{gu2025delta} on 20\% compressed Mixtral 6x7B. 
\textbf{Middle}: Effect of calibration sample size on perplexity.
\textbf{Right}: Trade-off between expert count, memory usage, and accuracy for Mixtral MoE.}
\label{fig:exp}
\vspace{-5mm}
\end{figure}

\textbf{Ablation on Core Components:}
Table \ref{tab:ablation_components} presents a comprehensive ablation study on key components of our Sub-MoE. 
For the Clustering component (A), we investigate three critical design choices:

\textbf{(1) Multi-layer  configuration in Adaptive Allocation} impacts performance, with 2-layer clustering (grouping 8×2=16 experts) yielding lowest perplexity and  highest average accuracy. This balanced approach provides sufficient flexibility for identifying functional relationships while maintaining manageable cluster sizes. In contrast, 1-layer clustering limits the diversity of potential merge candidates, while 3-layer clustering creates overly complex groupings that lead to accuracy drops.

\textbf{(2) Similarity Metric} comparison reveals that while router-logits and weight-based similarity measures perform reasonably well, our expert output similarity metric achieves the best overall balance between language modeling and reasoning tasks (0.64 mean accuracy). 
\textbf{(3) Clustering Algorithm} analysis shows that K-means consistently delivers optimal results compared to random grouping or hierarchical clustering, particularly on language modeling tasks, though hierarchical clustering achieves comparable performance on reasoning tasks.

For Merging component (B), we examine two key aspects:

\textbf{(5) $U$-Sharing strategy} comparison demonstrates that our union SVD approach substantially outperforms vanilla SVD across all metrics (8.54 vs. 10.67 on C4; 0.64 vs. 0.58 average accuracy), with particularly notable improvements on BoolQ (0.86 vs. 0.61). This confirms the effectiveness of our approach in finding a common representational space that preserves expert functionality.

\textbf{(6) $V$-Merging strategy} experiments show that our frequency-based approach consistently outperforms both dropping the least significant components and simple averaging. The frequency-weighted approach maintains better overall performance, demonstrating the importance of respecting expert utilization patterns when merging.
These ablation results empirically validate our design choices and demonstrate that each component of  Sub-MoE contributes meaningfully to its overall effectiveness.

\textbf{Impact of Calibration  Size .} Figure~\ref{fig:exp} (Middle) shows how calibration sample size affects model performance. Increasing samples from 32 to 128 substantially reduces perplexity on WikiText-2 (8.69→5.16) and C4 (23.89→8.54), while further increases yield minimal gains.

\textbf{Memory and Runtime Analysis.}
As shown in Figure~\ref{fig:exp} (Right), reducing Mixtral-8×7B from 8 to 6 experts decreases memory by 24\% with only a 6\% accuracy drop, while compression to 4 experts achieves optimal efficiency with 48\% memory reduction and 13\% accuracy decline. Compression below 4 experts causes disproportionate performance degradation, indicating a practical lower bound for maintaining capabilities. 
For runtime, our Sub-MoE$\dag$ can achieve 1.1$\sim$1.3× throughput speedup (from 87.7 to 120.9 tokens/second in Table \ref{tab:svd_ratio}) by compressing the weights of activated experts. 

\section{Conclusions}
In this paper, we present Sub-MoE, a new expert merging framework that addresses parameter conflicts in MoE LLM compression through subspace alignment. By decomposing concatenated experts via SVD, our approach extracts shared $U$-matrices while enabling the effective merging of expert-specific $V$ components. Our two-phase, Adaptive Expert Clustering and Subspace Expert Merging, identifies functionally similar experts and combines them with minimal information loss. Extensive experiments on Mixtral, DeepSeek, and Qwen MoE LLMs reveal that our approach consistently outperforms state-of-the-art pruning and merging baselines, achieving higher compression ratios with minimal loss in model efficacy. This superior performance stems from our ability to minimize parameter conflicts by operating in a common representation space and weighting expert contributions based on their activation patterns. Our approach offers immediate practical benefits for deploying MoE models on resource-constrained devices while opening promising research directions, including applying subspace alignment to other models and developing more sophisticated merging strategies.

\textbf{Limitations.} Following most MoE compressions which rely on calibration datasets, our method needs calibration data during clustering and merging. We will explore data-free ways in future work.

\bibliography{main}
\bibliographystyle{plain}

\end{document}